\pdfoutput=1
\documentclass[11pt]{article}

\usepackage[final]{acl}

\usepackage{times}
\usepackage{latexsym}
\usepackage{multirow}
\usepackage{multicol}
\usepackage{tabularx}
\usepackage{subcaption}

\usepackage{amsmath}
\usepackage{amssymb}

\usepackage[T1]{fontenc}
\usepackage[utf8]{inputenc}

\usepackage{microtype}

\usepackage{inconsolata}

\usepackage{graphicx}
\usepackage{booktabs}
\usepackage{ragged2e}

\title{Rebalancing Token Importance in Language Models with TF-IDF Weighted Cross-Entropy Loss}

\author{
Zhijian Li$^{1}$ \quad Stefan Larson$^{2}$ \quad Kevin Leach$^{2}$ \\
$^{1}$University of Southern California, Los Angeles \\
$^{2}$Vanderbilt University, Nashville
}

\begin{document}
\maketitle
\begin{abstract}
Large language models are typically trained under uniform token weighting, which allows frequent and low-information tokens to dominate learning and can increase the tendency to memorize surface-level text spans.
To address this, we present an information-weighted cross-entropy loss that rescales token-level contributions using TF-IDF statistics, emphasizing semantically informative tokens while downweighting ubiquitous ones. Experiments on five decoder-only LLMs ranging from 1.1B to 13B parameters show consistent reductions in memorized substring length while preserving perplexity and downstream task performance. Under LoRA fine-tuning, TF-IDF reduces average substring memorization length by 14\% across all five models. Under full-weight fine-tuning on TinyLLaMA~1.1B, the reduction reaches 58\%.
Our approach is architecture-agnostic and can be incorporated into existing training pipelines with less than 3\% computational overhead, offering a lightweight and principled way to mitigate memorization without disrupting standard training dynamics.

\end{abstract}

\section{Introduction}
Large language models (LLMs) are increasingly deployed in domains where reliability (e.g., clinical decision support), robustness (e.g., legal document analysis), and data stewardship (e.g., copyright-sensitive content generation) are essential.
Yet, their training objectives still treat all tokens equally, regardless of how informative, redundant, or noisy those tokens may be~\cite{touvron2023llama, brown2020language}.
This uniform weighting can misallocate gradient updates, leading to overemphasizing frequent or low-information patterns while underrepresenting distinctive or semantically-rich content~\cite{su2023mile}.
This can lead to memorization of surface-level patterns in the training data~\cite{carlini2022quantifying}. When every token is weighted equally, the model has no incentive to treat repeated surface patterns differently from meaningful content, leaving it prone to memorizing sequences verbatim.

Existing interventions include pre-training data deduplication~\cite{kandpal2022deduplicating}, which requires costly global pre-computation, and privacy-preserving objectives such as differential privacy~\cite{abadi2016deep}, which trade off model utility. Post-hoc methods such as unlearning~\cite{eldan2023whos} and model editing~\cite{meng2022locating} operate after training but do not address how the model learns during training. Token-dropping methods~\cite{hans2024like} offer a lighter-weight alternative but remove supervision on a subset of tokens rather than reweighting all of them. We instead focus on reshaping how the gradient signal is distributed across tokens during training. Our method is designed for the continued pretraining and fine-tuning regime, the dominant paradigm in modern NLP where practitioners adapt publicly released pretrained checkpoints rather than training from scratch~\citep{bommasani2021opportunities, howard2018universal}.

We present a TF-IDF-weighted cross-entropy loss that rescales token-level gradients using lexical statistics from the training corpus.  
The method up-weights tokens that are distinctive within context and down-weights those that are globally ubiquitous.  
Meanwhile, we retain supervision for all tokens, preserving coherent sequence modeling while reducing the incentive to memorize exact surface forms.  
The design is straightforward, architecture-agnostic, and readily integrates into existing training pipelines.

We evaluate this technique across five pretrained decoder-only LLMs ranging from 1.1B to 13B parameters. Empirically, the TF-IDF-weighted loss preserves perplexity and downstream performance on summarization and question answering (QA), while consistently reducing substring-level memorization and ROUGE-L across all models. The magnitude of the reduction is driven by how much memorization occurs under the CE baseline. Smaller models fine-tuned with LoRA already memorize little, leaving less room for improvement in that setting. Under full-weight fine-tuning, where the CE baseline memorizes substantially more, even the smallest model --- TinyLLaMA~1.1B --- achieves the largest reduction of 58\% in average Longest Memorized Substring (LMS).
Across all five LoRA models the average reduction is 14\%, indicating that TF-IDF loss consistently reduces memorization wherever the baseline permits it. Together, these results demonstrate that token-aware objectives provide an efficient and principled means of mitigating memorization without altering model architectures, standard training procedures, or downstream task performance.

\section{Related Work}

We review three areas of related work that inform and contextualize our approach: memorization in LLMs and existing mitigation strategies, regularization and alternative training objectives, and token-weighted loss functions.

\subsection{Memorization and Generalization in Large Language Models}
LLMs are known to verbatim memorize portions of their training data~\cite{carlini2019secret, carlini2021extracting}. 
Prior research has established various metrics to quantify this phenomenon, most notably \textit{exposure} and \textit{Longest Memorized Substring (LMS)}~\cite{yeom2018privacy, kandpal2022deduplicating}. 
Recent controlled studies suggest that such memorization is not merely a byproduct of model capacity, but is closely tied to the frequency of sequence repetition and the concentration of gradient updates on specific spans during training~\cite{huang2024demystifying}. 

Current mitigation strategies typically operate at the data level through deduplication and filtering~\cite{kandpal2022deduplicating}, at the optimization level via differential privacy~\cite{abadi2016deep}, or through post-hoc model unlearning~\cite{eldan2023whos} and model editing~\cite{meng2022locating}.
However, these techniques often force a trade-off: they either require massive pre-computation (deduplication) or dramatically degrade the model's downstream utility and reasoning capabilities.
A separate line of work modifies the training objective directly: token-dropping approaches~\cite{hans2024like} randomly exclude subsets of tokens from the loss to prevent the model from completing memorized chains, achieving strong reductions particularly at pre-training scale.
Our method occupies a different point in this design space --- rather than removing supervision on any token, it reweights all tokens using corpus statistics, preserving full-sequence modeling while redirecting gradient pressure away from common, low-information tokens.

\subsection{Regularization and Alternative Training Objectives}
Parallel to data-centric approaches, a broad line of work seeks to mitigate overfitting by constraining model capacity or softening the training objective. Classical techniques such as weight decay and dropout~\cite{krogh1991simple, nitish2014dropout} have been adapted for large transformers through methods like mixout or stochastic depth~\cite{lee2019mixout, huang2016deep}. Additionally, entropy regularization~\cite{pereyra2017regularizing} and label smoothing~\cite{muller2019label} discourage overconfidence by preventing the model from assigning total probability mass to a single token. 

While effective at improving general robustness, these methods operate primarily at the model or sequence level. They treat the loss landscape as uniform, without accounting for the fact that some tokens are inherently more prone to memorization than others. While regularization constrains \textit{how} a model learns, our method rebalances \textit{what} the model prioritizes in learning.

\subsection{Token-Weighted and Reweighted Loss Functions}
Recent work has explored modifying the loss function to assign non-uniform importance across tokens or examples. 
Focal loss~\cite{lin2017focal} down-weights easy examples in classification tasks to address class imbalance.  
MiLe Loss~\cite{su2023mile} reweights tokens according to the model's prediction entropy, emphasizing uncertain (hard-to-learn) tokens.  
Selective Language Modeling~\cite{lin2024not} computes token-utility scores using a reference model and trains only on high-utility tokens, improving data efficiency.  
Bilevel and meta-reweighting methods~\cite{ren2018learning,pan2024scalebio} learn example weights dynamically to optimize downstream validation performance.  
Together, these works demonstrate the growing interest in weighting strategies that better align gradient updates with token informativeness.

In contrast to these methods, which primarily rely on dynamic model-dependent metrics or auxiliary reference models, our TF-IDF objective utilizes static corpus statistics to establish token importance.
This provides a computationally efficient and linguistically principled alternative that targets the inherent informational density of language without the overhead of secondary models or iterative meta-optimization.
To our knowledge, applying TF-IDF statistics as a token-level loss weighting scheme for language model training has not been previously explored.
%FIXME you might want to say not just "X has not been done before" but also "We do X by leveraging P Q R"

\begin{figure*}[t]
    \centering
    \includegraphics[width=0.85\textwidth]{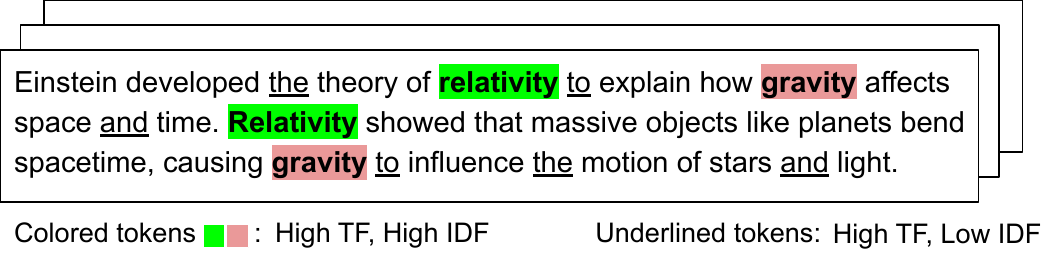}
    \caption{Illustration of TF-IDF weights applied to a short paragraph. Informative tokens such as \textit{relativity}, and  \textit{gravity} receive higher weights, while common words like  \textit{the} and \textit{and} are de-emphasized.}
    \label{fig:tfidf_example}
\end{figure*}

\section{TF-IDF Weighted Loss}

We present a \textbf{TF-IDF-weighted cross-entropy} objective that retains supervision on every token but rescales each term by a token-specific weight $w_i$:
\begin{equation}
\mathcal{L}_{\text{TF-IDF}}(\theta) = -\frac{1}{L}\sum_{i=1}^{L} w_i \log P_\theta(x_i \mid x_{<i})
\end{equation}

In contrast, standard autoregressive language models minimize cross-entropy (CE) loss over a token sequence $X = (x_1, \dots, x_L)$, corresponding to the special case $w_i = 1$:
\begin{equation}
\mathcal{L}(\theta) = -\frac{1}{L}\sum_{i=1}^{L}\log P_\theta(x_i \mid x_{<i})
\end{equation}

Computing IDF over the full training corpus --- which for LLMs can span billions of tokens --- would be prohibitively expensive; we instead maintain a rolling buffer of recent mini-batches to approximate corpus-level statistics efficiently. We then illustrate how these weights redirect gradient pressure toward semantically informative tokens and away from the common surface tokens that drive verbatim memorization.

\subsection{Buffer-Averaged TF-IDF Statistics}
Weights $w_i$ are computed using local TF and smoothed IDF estimated over an accumulation buffer of $K=16$ mini-batches ($N = B \times K$ sequences, where $B$ is the per-device batch size).
This buffer size balances re-weighting aggressiveness with statistical stability.
A larger $N$ widens the IDF gap between ubiquitous stop words ($df \approx N$) and rare keywords ($df=1$), shifting more gradient mass toward informative tokens.
We find $K=16$ provides a large enough window for robust frequency estimation while avoiding the overhead of global pre-computation.
The TF component further amplifies weight for tokens that repeat within a sequence, capturing locally salient terms whose within-context density signals domain relevance beyond what IDF alone conveys.
  
%FIXME if you're tight on space, I feel like TF-IDF is common enough that we could reduce these definitions. 
\paragraph{Term Frequency (TF).}
$\mathrm{tf}_i$ counts occurrences of token $x_i$ within its sequence:
\begin{equation}
\mathrm{tf}_i = \left| \{ j \in \{1, \dots, L\} : x_j = x_i \} \right|
\end{equation}

\paragraph{Document Frequency and IDF.}
Over the buffer of $N$ sequences, $\mathrm{df}(v)$ counts how many sequences contain token $v$, where $X^{(n)}$ denotes the $n$-th sequence and $\mathbb{I}(\cdot)$ is the indicator function. We apply smoothed IDF~\cite{schutze2008introduction} to keep weights positive even for near-universal tokens:
\begin{equation}
\mathrm{df}(v) = \sum_{n=1}^{N} \mathbb{I}(v \in X^{(n)})
\end{equation}
\begin{equation}
\mathrm{idf}(v) = \log \left( \frac{1+N}{1+\mathrm{df}(v)} \right) + 1
\end{equation}

\paragraph{Weight Normalization.}
Raw TF-IDF weights are computed as $w'_i = \mathrm{tf}_i \cdot \mathrm{idf}(x_i)$ and normalized to unit mean over the $M$ non-padding tokens in the mini-batch, ensuring $\mathbb{E}[w]=1$ so that standard learning rates remain applicable:
\begin{equation}
w_i = \frac{w'_i}{\frac{1}{M}\sum_{j=1}^{M} w'_j}
\end{equation}
Together, these components allow the model to internalize the relative importance of tokens during training, rebalancing gradient pressure in a way that reduces the incentive to memorize exact surface forms.

\subsection{Weighting Dynamics}

Consider a mini-batch drawn from diverse domains. In a sequence about scientific theory shown in Figure \ref{fig:tfidf_example}, tokens like \textit{relativity} and \textit{gravity} appear frequently within that context but rarely in other sequences, so they receive high TF-IDF weights and the model is pushed to predict them accurately. Function words like \textit{the} and \textit{and} appear in nearly every sequence and are down-weighted, shifting gradient mass toward semantically meaningful content.
In noisy datasets, common formatting symbols and boilerplate text appear across many sequences, giving them high document frequency and low IDF weight, similar to function words.

The TF-IDF-weighted objective retains all tokens in the loss, rebalancing where gradient pressure falls rather than removing supervision on any position. This discourages memorization because gradient signal is no longer uniformly spread across all positions. Pressure concentrates on tokens that are locally frequent and globally rare, the distinctive content words that TF-IDF upweights, rather than on the common surface tokens that define a sequence's exact wording.

\section{Experimental Setup}

\subsection{Models}
We assess five commonly used decoder-only LLMs spanning 1.1B--13B parameters: \textbf{TinyLLaMA~1.1B}~\citep{zhang2024tinyllama}, a compact LLaMA-style model; \textbf{Pythia~1.4B}~\citep{biderman2023pythia}, trained on The Pile with transparent checkpoints; \textbf{GPT-J~6B}~\citep{wang2021gpt}, a widely adopted open-source model also trained on The Pile; \textbf{LLaMA-2~7B}~\citep{touvron2023llama}, a strong mid-size foundation model; and \textbf{LLaMA-2~13B}~\citep{touvron2023llama}, its larger sibling. This range enables testing generalization of our method across model scales. We do not include smaller models, as they often produce outputs too short or simplistic for meaningful memorization and downstream evaluation.

\subsection{Fine-Tuning Setup}
Consistent with our focus on the continued pretraining and fine-tuning regime, all experiments initialize from publicly released pretrained checkpoints.
This reflects how LLM practitioners typically operate~\citep{bommasani2021opportunities, howard2018universal}: starting from a strong pretrained foundation and adapting it to new objectives or domains, rather than training from scratch.

For our primary experiments across the full model suite (1.1B--13B parameters), we use Low-Rank Adaptation (LoRA)~\cite{hu2022lora}, which adds trainable low-rank matrices to the attention projection layers while keeping the original weights frozen.
Prior work shows LoRA can match or exceed full fine-tuning on language modeling and downstream tasks~\citep{hu2022lora, dettmers2023qlora, lialin2023scaling}, and it remains practical at all model scales we consider.
We use rank $r{=}8$ and scaling factor $\alpha{=}32$ (memorization and QA) or $\alpha{=}16$ (summarization), with a learning rate of $1\times10^{-4}$ and the AdamW optimizer. Per-model details including target modules, quantization, and batch sizes are provided in Appendix~\ref{app:config}.

To verify that our findings are not artifacts of the LoRA parameterization, we also run full-weight fine-tuning on TinyLLaMA~1.1B, updating all model parameters. We use a reduced learning rate of $5\times10^{-5}$, as applying the same LoRA learning rate to full-weight training caused catastrophic memorization.

\subsection{Evaluation Categories and Datasets}\label{evaluation_setup}
We evaluate across four categories: \textit{Memorization}, \textit{Perplexity}, \textit{Summarization}, and \textit{QA}. For each model, training and decoding protocols are identical across conditions and only the loss function differs. All generation-based evaluations use deterministic greedy decoding.

\begin{table*}[t]
\centering
\small
\setlength{\tabcolsep}{6pt}
\renewcommand{\arraystretch}{1.0}
\begin{tabular*}{\textwidth}{@{\extracolsep{\fill}}lcccccc}
\toprule
\textbf{Model} & \textbf{Objective} & \textbf{Avg Prefix} & \textbf{Max Prefix} & \textbf{Avg LMS} & \textbf{Max LMS} & \textbf{ROUGE-L} \\
\midrule
\multirow{2}{*}{TinyLLaMA 1.1B$^\dagger$}
 & CE & 0.00 & 0 & 3.23 & 10 & 17.5 \\
 & TF-IDF & 0.00 & 0 & \textbf{3.22} & 10 & \textbf{17.1} \\
\midrule
\multirow{2}{*}{TinyLLaMA 1.1B$^\ddagger$}
 & CE & 5.65 & 128 & 8.30 & 128 & 21.05 \\
 & TF-IDF & \textbf{1.36} & \textbf{18} & \textbf{3.50} & \textbf{18} & \textbf{13.55} \\
\midrule
\multirow{2}{*}{Pythia 1.4B}
 & CE & 0.63 & 6 & 2.44 & 9 & 17.54 \\
 & TF-IDF & \textbf{0.57} & \textbf{5} & \textbf{2.07} & \textbf{7} & \textbf{17.31} \\
\midrule
\multirow{2}{*}{GPT-J 6B}
 & CE & 1.08 & 10 & 4.46 & 21 & 20.3 \\
 & TF-IDF & \textbf{0.00} & \textbf{0} & \textbf{3.55} & \textbf{10} & \textbf{19.2} \\
\midrule
\multirow{2}{*}{LLaMA-2 7B}
 & CE & 1.37 & 14 & 3.48 & 14 & 18.92 \\
 & TF-IDF & \textbf{1.02} & \textbf{7} & \textbf{2.98} & \textbf{10} & \textbf{17.68} \\
\midrule
\multirow{2}{*}{LLaMA-2 13B}
 & CE & 1.82 & 14 & 3.90 & 14 & 19.79 \\
 & TF-IDF & \textbf{1.17} & \textbf{11} & \textbf{3.17} & \textbf{11} & \textbf{18.40} \\
\bottomrule
\end{tabular*}
{\small $^\dagger$LoRA fine-tuning. $^\ddagger$Full-weight fine-tuning.}
\caption{\textbf{Memorization Metrics across Models and Objectives.}
Comparison between standard cross-entropy (CE) and TF-IDF-weighted cross-entropy.
For all reported metrics, \textbf{lower values} indicate superior mitigation of verbatim recall. LMS (Longest Memorized Substring) quantifies the length of the longest exact token substring recovered from the training set, while prefix matches represent exact recall triggered by the start of a sequence. All metrics are averaged across prefix lengths $\mathcal{P}=\{32, 50, 100\}$.}\label{tab:memorization_full}
\end{table*}

\paragraph{Memorization.}
To evaluate verbatim recall, we adopt the \textit{controlled injection} framework utilized by~\citet{huang2024demystifying} to study memorization in modern LLMs.
We construct a training corpus consisting of 20,000 base sequences from the Pile-uncopyrighted dataset~\citep{gao2020pile}, into which we inject 100 target sequences from WikiText-2~\citep{merity2016pointer}. Each target sequence is repeated 10 times at random positions to simulate data duplication. We initialize from pretrained checkpoints and train for one epoch using LoRA (batch size 8, lr $1\times10^{-4}$, block size 256). For full-weight fine-tuning on TinyLLaMA~1.1B, we use the same corpus and epoch count with all parameters updated (batch size 8, block size 256).

At evaluation time, we probe the model's recall using prefix lengths $\mathcal{P}=\{32,50,100\}$ tokens. For a given \(p\in\mathcal{P}\), the model is conditioned on the first \(p\) tokens and asked to continue. We evaluate checkpoints saved at \(10\%,25\%,50\%,75\%,100\%\) of training progress to observe how memorization evolves throughout training, not only at the final checkpoint.

For each (sequence, $p$) pair, we compare the generated continuation to the ground truth using three metrics, where lower values indicate less memorization:  (1) \textit{Longest prefix match} counts consecutive token matches from the start of the continuation; (2)  \textit{Longest Memorized Substring (LMS)}~\cite{huang2024demystifying} measures the longest exact token substring shared between the generation and the ground truth; and (3) \textit{ROUGE-L}~\cite{lin2004rouge} captures broader structural similarity.

\paragraph{Perplexity.}
We measure token-level perplexity on an unseen corpus using the same checkpoints from the memorization setup. Perplexity is evaluated on the \textit{WikiText-2 validation} split, which is disjoint from the injected training data. We tokenize the full validation set, concatenate tokens into non-overlapping blocks of 256, and compute:
\[
\mathrm{PPL} \;=\; \exp\!\left(\frac{1}{|\mathcal{D}|}\sum_{t\in\mathcal{D}} -\log P_\theta\big(x_t \,\big|\, x_{<t}\big)\right)
\]
where $\mathcal{D}$ indexes all evaluated token positions. This isolates generalization to unseen data while keeping the training setup identical across models and loss variants.

\paragraph{Summarization.}
We fine-tune models on the \textsc{CNN/DailyMail} v3.0.0 train split and evaluate on validation~\citep{hermann2015teaching}, using an instruction-style prompt (\texttt{``summarize: ''}) to elicit summary highlights. Models are trained for 1 epoch (batch size 4, lr $2\times10^{-4}$, max source length 1024, max target length 128). Quality is measured with \textit{ROUGE-1/2/L}~\citep{lin2004rouge} for n-gram overlap and \textit{BERTScore-F1}~\citep{zhang2019bertscore} for semantic similarity.

\paragraph{QA.}
We fine-tune on the \textsc{SQuAD} v1.1~\citep{rajpurkar2016squad} train split and evaluate on validation. Models receive a context passage and a question and must generate a short answer span. Models are trained for 1 epoch (batch size 8, lr $1\times10^{-4}$, block size 256). We report \textit{Exact Match} (EM), which requires character-level identity with the ground truth, and \textit{F1 Score}, which measures word-level overlap between the predicted and reference answers.

We select summarization and QA as downstream tasks because they represent complementary dimensions of language understanding: summarization tests the ability to produce coherent, abstractive output, while QA tests precise comprehension and span extraction. Together they provide broad coverage for assessing whether TF-IDF weighting preserves general utility.

\begin{figure*}[t]
    \centering
    % -------- Left subfigure --------
    \begin{subfigure}[t]{0.45\textwidth}
        \centering
        \includegraphics[width=\textwidth]{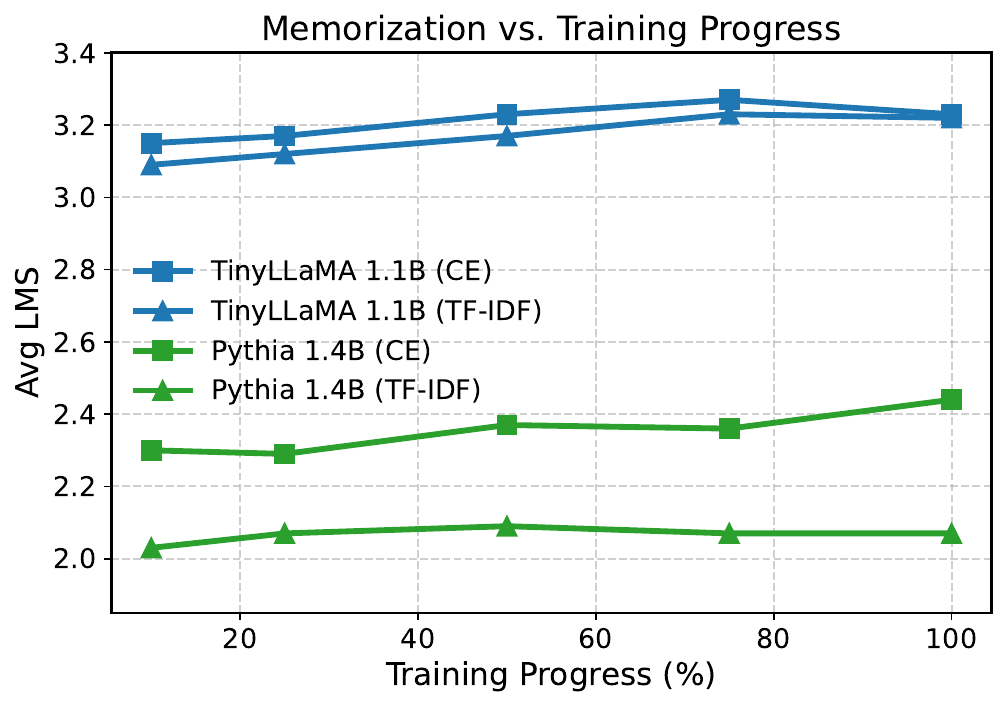}
        \caption{Smaller models: TinyLLaMA~1.1B and Pythia~1.4B.}
        \label{fig:memorization_small}
    \end{subfigure}
    \hfill
    % -------- Right subfigure --------
    \begin{subfigure}[t]{0.45\textwidth}
        \centering
        \includegraphics[width=\textwidth]{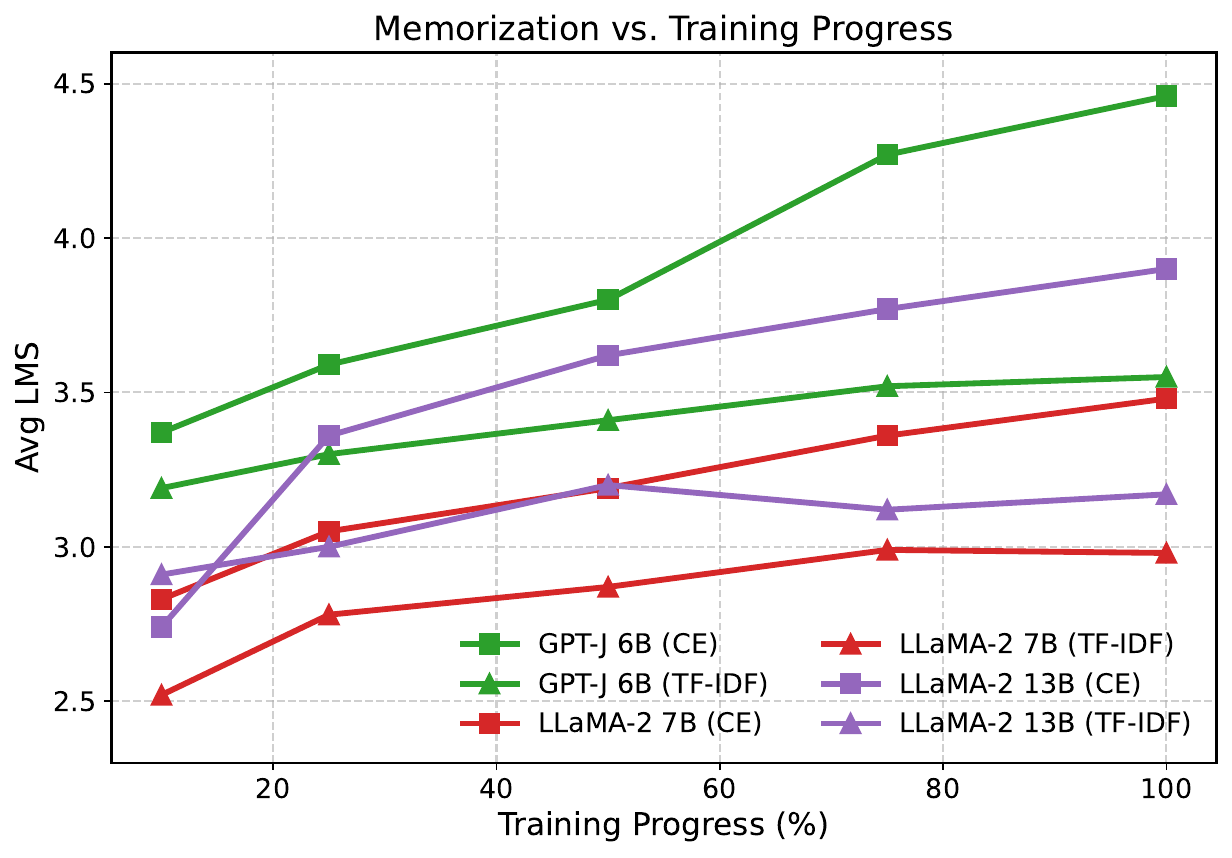}
        \caption{Larger models: GPT-J~6B and LLaMA-2~7B/13B.}
        \label{fig:memorization_large}
    \end{subfigure}
    
    % -------- Combined caption --------
    \caption{Avg LMS versus training progress under standard cross-entropy (CE) and TF-IDF-weighted objectives.
    Subfigure~(a) shows smaller models; subfigure~(b) shows mid- and large-scale models where the gap between standard CE and TF-IDF loss widens.}
    \label{fig:memorization_combined}
\end{figure*}

\section{Does TF-IDF-Weighted Loss Reduce Memorization?}
Table~\ref{tab:memorization_full} and Figure~\ref{fig:memorization_combined} summarize memorization behavior under CE and TF-IDF-weighted objectives. All metrics in this section --- Avg/Max Prefix, Avg/Max LMS, and ROUGE-L --- measure verbatim memorization; lower values indicate less memorization and better mitigation.
Under LoRA fine-tuning, no model reproduced entire passages verbatim; full-sequence matches were zero across all configurations.
Prefix matches under LoRA remain modest (averaging below 2 tokens across all models) though larger models show non-trivial values (e.g., LLaMA-2 13B CE: Avg Prefix = 1.82, Max = 14).
Full-weight fine-tuning on TinyLLaMA~1.1B is a more pronounced exception: CE training produces Avg Prefix = 5.65 with a maximum of 128 tokens, confirming that unconstrained parameter adaptation creates substantially more memorization pressure than LoRA. Prior work has shown that even short memorized spans can be sufficient to extract sensitive training data~\cite{carlini2021extracting}, motivating reduction even when absolute values appear modest.

Across all five LoRA models, the average reduction in LMS is 14\%. For the four LoRA-tuned models excluding TinyLLaMA, the reduction in memorization is consistent and substantial.
GPT-J~6B's average LMS falls by over 20\%, from 4.46 to 3.55, while its maximum span is halved from 21 to 10 tokens.
LLaMA-2~7B and 13B show reductions in both average LMS (3.48 to 2.98 and 3.90 to 3.17) and maximum span (from 14 to 10 and 14 to 11, respectively).
Pythia~1.4B drops from 2.44 to 2.07.
ROUGE-L follows the same trend across all four models.
Percentile bootstrap CIs ($n{=}10{,}000$ resamples, 95\%) confirm statistically significant reductions for Pythia~1.4B (15.0\%, CI [6.9\%, 22.6\%]), LLaMA-2~7B (14.4\%, CI [5.6\%, 22.4\%]), and LLaMA-2~13B (18.6\%, CI [10.4\%, 25.9\%]), all excluding zero.
As shown in Figure~\ref{fig:memorization_combined}, TF-IDF training produces lower LMS at every training checkpoint, introducing a stable downward offset without changing the rate at which memorization accumulates.
This gap widens for larger models as training progresses.

TinyLLaMA~1.1B under LoRA shows a negligible difference (3.23 to 3.22), with a bootstrap CI on the relative reduction of [$-$8.3\%, 8.3\%] that includes zero, confirming this result is not statistically significant.
We attribute this to LoRA's parameter constraints already suppressing memorization to a level that is equally low under both objectives, leaving little room for TF-IDF to further reduce it.
Full-weight fine-tuning directly tests this: without LoRA, CE training drives average LMS to 8.30 with a maximum span of 128 tokens, far higher than the LoRA counterpart.
TF-IDF then produces a 58\% reduction in average LMS (to 3.50), cutting the maximum span to 18 tokens and ROUGE-L from 21.05 to 13.55.
Bootstrap confidence intervals ($n{=}10{,}000$ resamples, 95\%) confirm this reduction is statistically significant: the relative reduction of 57.8\% has a CI of [34.9\%, 71.1\%], which excludes zero.
Notably, the CE interval [5.45, 11.94] is substantially wider than the TF-IDF interval [3.18, 3.86], indicating that TF-IDF not only reduces memorization on average but also makes it more consistent across sequences.
This confirms that TF-IDF's effect scales with the model's actual capacity to memorize.
When LoRA constrains the parameter space, both objectives reach similarly low memorization.
When all parameters are free, gradient rebalancing provides strong protection.

Table~\ref{tab:qualitative} illustrates this contrast concretely.
Given a 32-token prefix from an injected WikiText-2 sequence, the CE model reproduces the continuation verbatim (LMS=128, ROUGE-L=100).
The TF-IDF model generates a topically related but clearly non-memorized continuation (LMS=2, ROUGE-L=17.1).

\begin{table}[t]
\centering
\small
\begin{tabular}{p{1.5cm} p{5.5cm}}
\toprule
\textbf{} & \textbf{Text} \\
\midrule
Prompt & \textit{For several years the arsenal, which was owned by the federal government, served as a simple arms depot and was staffed with only a handful of soldiers\ldots} \\
\midrule
Ground truth & But in November 1860, with the American Civil War on the horizon, a company of the Second United States Artillery, consisting of sixty-five men, was transferred to Little Rock\ldots \\
\midrule
CE \newline {\small LMS=128, R-L=100} & But in November 1860, with the American Civil War on the horizon, a company of the Second United States Artillery, consisting of sixty-five men, was transferred to Little Rock\ldots \\
\midrule
TF-IDF \newline {\small LMS=2, R-L=17.1} & By the time the arsenal was officially closed in 1917, the United States had a military force of over 100,000 men\ldots \\
\bottomrule
\end{tabular}
\caption{Qualitative example: CE reproduces the injected sequence verbatim while TF-IDF generates a related but non-memorized continuation. Full-weight fine-tuning, prefix length 32; 128 tokens generated. R-L = ROUGE-L.}\label{tab:qualitative}
\end{table}

\begin{table}[t]
\centering
\small
\begin{tabular}{lcc}
\toprule
\textbf{Model} & \textbf{CE} $\downarrow$ & \textbf{TF-IDF} $\downarrow$ \\
\midrule
TinyLLaMA~1.1B$^\dagger$   & 25.58 & \textbf{23.59} \\
TinyLLaMA~1.1B$^\ddagger$  & 20.94 & \textbf{16.06} \\
Pythia~1.4B       & \textbf{17.23} & 18.36 \\
GPT-J~6B          & 18.80 & \textbf{17.59} \\
LLaMA-2~7B        & 10.86 & \textbf{10.65} \\
LLaMA-2~13B       & 10.35 & \textbf{10.28} \\
\bottomrule
\multicolumn{3}{l}{\small $^\dagger$LoRA fine-tuning. $^\ddagger$Full-weight fine-tuning.}
\end{tabular}
\caption{Perplexity on the WikiText-2 validation set for models fine-tuned with standard cross-entropy (CE) and TF-IDF-weighted objectives. Lower is better.}\label{tab:perplexity}
\end{table}

\begin{table*}[t]
\centering
\small
\begin{tabular*}{\textwidth}{@{\extracolsep{\fill}}lcccccccc}
\toprule
\multirow{2}{*}{\textbf{Model}} &
\multicolumn{2}{c}{\textbf{ROUGE-1}} &
\multicolumn{2}{c}{\textbf{ROUGE-2}} &
\multicolumn{2}{c}{\textbf{ROUGE-L}} &
\multicolumn{2}{c}{\textbf{BERTScore-F1}} \\
\cmidrule(lr){2-3} \cmidrule(lr){4-5} \cmidrule(lr){6-7} \cmidrule(lr){8-9}
 & CE & TF-IDF & CE & TF-IDF & CE & TF-IDF & CE & TF-IDF \\
\midrule
TinyLLaMA~1.1B$^\dagger$   & \textbf{32.4} & 31.9 & \textbf{13.5} & 13.3 & \textbf{24.3} & 24.0 & \textbf{88.0} & 87.6 \\
TinyLLaMA~1.1B$^\ddagger$  & \textbf{33.6} & 33.4 & \textbf{14.1} & 14.0 & \textbf{25.0} & 25.0 & \textbf{88.7} & 88.6 \\
Pythia~1.4B     & 28.5 & \textbf{28.7} & 16.2 & \textbf{16.3} & 20.7 & \textbf{20.8} & 84.8 & 84.8 \\
GPT-J~6B        & 28.5 & \textbf{29.0} & 16.2 & \textbf{16.5} & 20.7 & \textbf{20.9} & 84.9 & 84.9 \\
LLaMA-2~7B      & 32.9 & \textbf{33.0} & 18.6 & 18.6 & 23.9 & 23.9 & 84.4 & 84.4 \\
LLaMA-2~13B     & \textbf{28.3} & 28.1 & \textbf{15.0} & 14.8 & 20.3 & 20.3 & 83.9 & 83.9 \\
\bottomrule
\end{tabular*}
{\small $^\dagger$LoRA. $^\ddagger$Full-weight.}
\caption{Summarization performance on the \textsc{CNN/DailyMail} validation set under standard cross-entropy (CE) and TF-IDF-weighted objectives. Higher is better.}\label{tab:summarization}
\end{table*}

\begin{table*}[t]
\centering
\small
\begin{tabular*}{\textwidth}{@{\extracolsep{\fill}}lcccc}
\toprule
\textbf{Model} & \multicolumn{2}{c}{\textbf{Exact Match (\%)}} & \multicolumn{2}{c}{\textbf{F1 (\%)}} \\
\cmidrule(lr){2-3} \cmidrule(lr){4-5}
 & CE & TF-IDF & CE & TF-IDF \\
\midrule
TinyLLaMA~1.1B$^\dagger$   & \textbf{24.63} & 23.91 & 47.88 & \textbf{48.38} \\
TinyLLaMA~1.1B$^\ddagger$  & \textbf{79.00} & 78.40 & \textbf{84.46} & 84.18 \\
Pythia~1.4B     & \textbf{54.59} & 54.26 & 73.92 & \textbf{74.24} \\
GPT-J~6B        & \textbf{57.51} & 56.48 & \textbf{77.81} & 77.40 \\
LLaMA-2~7B      & \textbf{88.61} & 87.46 & \textbf{94.41} & 94.10 \\
LLaMA-2~13B     & \textbf{89.33} & 87.92 & \textbf{95.04} & 94.52 \\
\bottomrule
\end{tabular*}
{\small $^\dagger$LoRA. $^\ddagger$Full-weight.}
\caption{QA performance on the \textsc{SQuAD v1.1} validation set under standard cross-entropy (CE) and TF-IDF-weighted objectives. Higher is better.}\label{tab:qa}
\end{table*}

\section{Impact on Language Modeling and Downstream Performance}

Having shown that TF-IDF weighting reduces memorization, we now assess whether it preserves model utility across three dimensions: perplexity, summarization, and QA.

\subsection{Perplexity}
Lower perplexity indicates better generalization to unseen text. Table~\ref{tab:perplexity} shows that TF-IDF weighting achieves comparable or lower perplexity than CE across most models. We note that training data includes WikiText-2 injections while perplexity is measured on the WikiText-2 validation split; although these splits are disjoint, the domain overlap means perplexity differences partly reflect how each objective weights WikiText-2-style tokens rather than purely out-of-domain generalization. GPT-J~6B improves from 18.80 to 17.59, TinyLLaMA~1.1B from 25.58 to 23.59, and both LLaMA-2 models see small gains as well. Only Pythia~1.4B records a minor increase (17.23 to 18.36), the sole exception across all six configurations. Overall, TF-IDF reweighting does not degrade language modeling quality; perplexity remains stable or slightly improved across all architectures.

Under full-weight training, the improvement is more pronounced: TF-IDF achieves 16.06 versus 20.94 for CE on TinyLLaMA~1.1B, suggesting that emphasizing informative tokens provides a stronger regularization signal when all parameters are free to adapt.

\subsection{Summarization}

Unlike the memorization metrics above, all downstream scores in this section and the following QA section are higher-is-better, reflecting model utility rather than verbatim recall. Table~\ref{tab:summarization} shows that summarization performance is stable across all models under TF-IDF weighting. Scores are nearly indistinguishable from the CE baseline in most cases. LLaMA-2~7B achieves near-identical results (33.0 vs. 32.9 ROUGE-1), while Pythia~1.4B and GPT-J~6B show small gains. TinyLLaMA~1.1B and LLaMA-2~13B show fluctuations of around 0.2--0.5 points, which are negligible in the context of overall generation quality. For full-weight TinyLLaMA~1.1B, the CE baseline and TF-IDF also yield comparable scores across all metrics. Although ROUGE is a limited proxy for abstractive quality~\cite{bhandari2020re}, the agreement between ROUGE and BERTScore-F1 across all models suggests that TF-IDF weighting does not disrupt generation quality for summarization tasks.

\subsection{QA}
Table~\ref{tab:qa} shows that TF-IDF weighting preserves QA performance across all model scales. Differences in Exact Match (EM) and F1 are within 1.5\% in all cases. TinyLLaMA~1.1B shows a small drop in EM but a small gain in F1, while Pythia~1.4B is nearly identical under both settings. LLaMA-2~13B shows the largest EM drop ($-$1.41\%), the closest to the 1.5\% threshold; F1 remains high at 94.52 versus 95.04 under CE.

Under full-weight fine-tuning, TinyLLaMA~1.1B reaches EM/F1 of 79.0/84.46 under CE and 78.4/84.18 under TF-IDF, a gap of less than 1\%. The much higher absolute scores compared to LoRA (CE EM=24.63) reflect the greater capacity of full-weight fine-tuning for task adaptation, while the near-identical CE and TF-IDF results confirm that the objective does not impair extractive reasoning.

\paragraph{Summary.} Across all three evaluation dimensions, TF-IDF weighting produces results within noise of the CE baseline. Perplexity improves for five of six model configurations. Summarization scores diverge by at most 0.5 ROUGE points. QA differences are within 1.5\% EM and 0.6\% F1. Together, these results confirm that consistent memorization reduction comes at no meaningful cost to downstream utility.

\section{Conclusion}
In this work, we address the phenomenon of verbatim memorization in LLMs by challenging the standard practice of uniform token weighting during training. We present a \textit{TF-IDF-weighted cross-entropy objective} that rebalances the learning signal according to lexical information density, effectively prioritizing semantically rich tokens over high-frequency, low-entropy ones.

Our experiments across five decoder-only LLMs, ranging from 1.1B to 13B parameters, demonstrate that this re-weighting consistently mitigates memorization---specifically reducing the Longest Memorized Substring and ROUGE-L---without compromising linguistic fluency or downstream performance on summarization and QA tasks, while tracking memorization across training checkpoints confirms a stable downward shift rather than a mere delay in onset.

The TF-IDF-weighted objective is architecture-agnostic and introduces less than 3\% computational overhead (256\,ms vs.\ 262\,ms per step on TinyLLaMA~1.1B, batch size 8, full-weight), making it directly applicable to standard training pipelines. These findings highlight the potential of token-aware objectives as a scalable, lightweight strategy for mitigating memorization in service of privacy and data stewardship. Future directions include extending evaluation to adversarial extraction, mechanistic analysis of lexical reweighting, and testing TF-IDF objectives during pretraining and domain adaptation.

\section*{Limitations}

Our work has several limitations that provide context for our findings and suggest directions for future research.

\textbf{Evaluation scope.} Our memorization evaluation focuses on verbatim exact-match recall following established practice~\citep{carlini2019secret, huang2024demystifying}, as this represents the most direct risk for privacy and copyright. It does not cover semantic paraphrasing or adversarial prompting techniques designed to extract training data.

\textbf{Training regime.} While most experiments use LoRA fine-tuning, we validate the method under full-weight training on TinyLLaMA~1.1B. Neither setting captures how TF-IDF weighting would interact with training from scratch, where consistently down-weighting function words could potentially hinder early acquisition of grammatical structure.

\textbf{Subword tokenization.} Classical TF-IDF operates at the word level, but our method applies weights to subword tokens produced by BPE tokenizers. A rare word may be split into subword units that are individually common, potentially diluting the intended upweighting of informative content. Analyzing the empirical weight distribution over subword tokens is a useful direction for future work.

\textbf{Context length.} Hardware constraints limited our block size to 256 tokens. This is sufficient for common memorization patterns but may not capture long-range memorization dynamics in models with larger context windows.

\section*{Ethics Statement}
This work adheres to the ACL Ethics Policy. Our experiments utilize publicly available datasets (CNN/DailyMail, SQuAD, and WikiText-2) in accordance with their intended research use. By proposing a loss function that reduces verbatim memorization, this research aims to enhance the privacy and safety of LLMs.

\section*{AI Usage Disclosure}
The authors used ChatGPT (OpenAI) and Claude (Anthropic) to assist in the linguistic polishing and grammatical refinement of this manuscript to improve clarity and readability. Additionally, these tools were used to assist in debugging the custom Python scripts used for the TF-IDF weighted loss implementation and the memorization evaluation pipeline. The core research objectives, the mathematical derivation of the buffer-averaged TF-IDF loss function, the experimental design, and the interpretation of all results were performed solely by the human authors.

\bibliography{custom}

\appendix

\section{Per-Model Training Configuration}
\label{app:config}

Table~\ref{tab:model_config} lists the full LoRA configuration used for each model in the memorization experiments. All models use rank $r{=}8$, $\alpha{=}32$, learning rate $1\times10^{-4}$, block size 256, 1 epoch, and the AdamW optimizer.

\begin{table}[h]
\centering
\small
\begin{tabular}{lccc}
\toprule
\textbf{Model} & \textbf{Batch} & \textbf{4-bit} & \textbf{dtype} \\
\midrule
TinyLLaMA~1.1B & 8 & No  & bf16 \\
Pythia~1.4B    & 8 & Yes & bf16 \\
GPT-J~6B       & 8 & No  & fp16 \\
LLaMA-2~7B     & 8 & Yes & bf16 \\
LLaMA-2~13B    & 8 & Yes & bf16 \\
\bottomrule
\end{tabular}
\caption{Per-model LoRA configuration. 4-bit uses NF4 with bfloat16 compute dtype.}\label{tab:model_config}
\end{table}

\begin{table}[h]
\centering
\small
\begin{tabular}{ll}
\toprule
\textbf{Model} & \textbf{LoRA Target Modules} \\
\midrule
TinyLLaMA~1.1B & \texttt{q\_proj, k\_proj, v\_proj, o\_proj} \\
Pythia~1.4B    & \texttt{query\_key\_value} \\
GPT-J~6B       & \texttt{attn.q\_proj, attn.k\_proj,} \\
               & \texttt{attn.v\_proj, attn.out\_proj} \\
LLaMA-2~7B     & \texttt{q\_proj, k\_proj, v\_proj, o\_proj} \\
LLaMA-2~13B    & \texttt{q\_proj, k\_proj, v\_proj, o\_proj} \\
\bottomrule
\end{tabular}
\caption{LoRA target modules per model architecture.}\label{tab:lora_modules}
\end{table}

\section*{Code and Data Availability}
The implementation of the TF-IDF Weighted Trainer and experimental scripts are available at \href{https://github.com/anonymoussubmission25/Rebalancing-Token-Importance-in-Language-Models-with-TF-IDF-Weighted-Cross-Entropy-Loss}{our GitHub repository}.

\end{document}